# Method and Dataset Mining in Scientific Papers

Rujing Yao*, Linlin Hou*, Yingchun Ye*,
Ou Wu
CAM, Tianjin University, China
{rjyao, ycye, wuou}@tju.edu.cn

Ji Zhang
Insitute of AI, Zhejiang Lab, Hangzhou,
China
zhangji77@gmail.com

Jian Wu
LAMPS Lab, Old Dominion University,
USA
jwu@cs.odu.edu

*Abstract*—Literature analysis facilitates researchers better understanding the development of science and technology. The conventional literature analysis focuses on the topics, authors, abstracts, keywords, references, etc., and rarely pays attention to the content of papers. In the field of machine learning, the involved methods (M) and datasets (D) are key information in papers. The extraction and mining of M and D are useful for discipline analysis and algorithm recommendation. In this paper, we propose a novel entity recognition model, called MDER, and constructe datasets from the papers of the PAKDD conferences (2009-2019). Some preliminary experiments are conducted to assess the extraction performance and the mining results are visualized.

## I. INTRODUCTION

Literature analysis is important. It can assist researchers to understand the main information of papers in a convenient way. Conventional literature analysis methods include the abstracts analysis [1], keywords extraction [2], the analysis of the cooperation relationship among authors [3], etc. In the machine learning literature, we find that the methods (M) and datasets (D) used in the experimental part are important, but researchers pay attention to them rarely. M and D provide readers with more detailed academic portraits of the papers, which can complement the conventional literature analysis. They can reflect the relationship between the metadata M and D and the development trends.

Therefore, the paper investigates a relatively new data mining problem which focuses on the involved methods (both the proposed and competing) and datasets in machine learning and data mining papers. The mining results are particularly useful in two aspects 1) M and D-based scientometrics in machine learning and data mining, which provide a supplementary to existing techniques based on paper meta data such as authors, keywords; and 2) algorithm recommendation if the relationships between the existing M and D records are well established. The primary challenge is the M and D extraction.

In this paper, we deal the problem as an entity recognition task in NLP [4, 5]. We propose a novel model called MDER on the basic of classical sequence labeling [6, 7, 8] and construct two new datasets. We use MDER to extract entities in papers of the PAKDD conferences from 2009 to 2019, analyze the relationship among different methods, and construct complex network graphs and the histogram of the betweenness centrality.

## II. METHODOLOGY

Problem description: For a given sentence $\{w_1, w_2, \cdots, w_n\}$, where $w_i$ represents each word, our purpose is to identify the label of each character in words, corresponding to M, D and others [9, 10]. Character-level labeling can almost avoid the appearance of new words. Our model combines the Rule-based technique and a new deep network structure, namely, CNN-BiLSTM+Attention+CRF. The structure of the model is shown in Fig. 1. Specifically, we construct one blacklist containing some general words (e.g. "the") and two whitelists containing the two entities. The labels of each character in the blacklist and whitelists are regarded as additional supervised information to aid model training. The method whitelist contains some common M entities, such as: "SVM", "LSTM". The labels of each character in these words are {B-M, I-M}. The dataset whitelist contains some D entities, such as: "Douban", "Twitter", so the labels of each character in these words are {B-D, I-D}. The blacklist contains some general words, such as: "the", "are", so the labels of each character in these words are {O}. Characters of the words which do not belong to the blacklist or whitelists are set to <unk>. For an input sentence, each character adopts rule embedding according to the above rules. Then, the rule embedding is concatenated with the character embedding as the input of both CNN and 2-layer BiLSTM. After the output vectors of these two modules are concatenated, the attention mechanism and the conditional random field (CRF) are used.

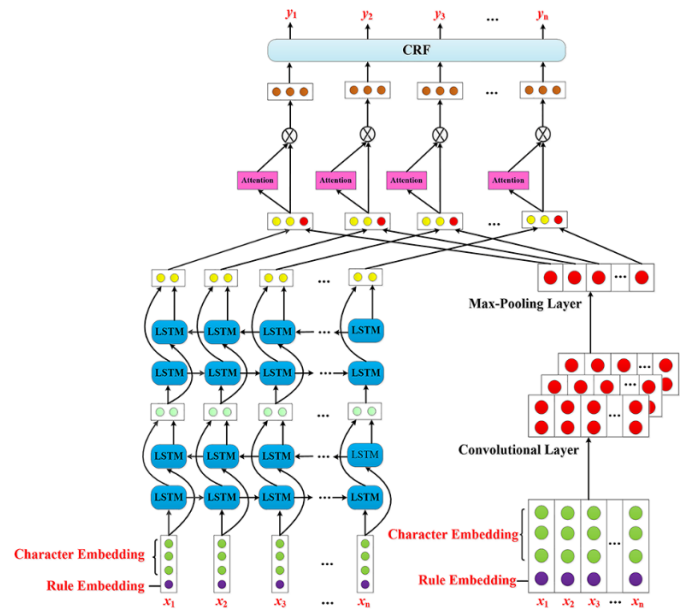

Fig. 1. The structure of MDER.

This work is supported by the Zhejiang Lab Fund (2019KB0AB03), Tianjin Nature Science Fund (19JCZDJC31300), and NSFC (61673377). Linlin Hou is a student at Nankai University and internship at Tianjin University currently. The first three authors contributed equally. Ou Wu is the corresponding author.

## III. EXPERIMENTS

### A. MDdata

Existing entity recognition datasets are inappropriate for training and evaluating MDER. Thus, two new datasets, MDdata1 and MDdata2, are constructed. The construction process of MDdata1 is as follows: Firstly, 430 papers of PAKDD conferences (2017-2019) and 266 papers of ACL 2019 are collected. We extract experiment chapters of these papers and divide them into 6,009 sentences. Secondly, we recruit six graduate students in our institute to label these sentences. The sentences are labeled with a standard process. Every two people mark the same sentences, and when the labeling coincidence rate is more than 95%, the labels are considered to be effective. Finally, considering that the number of M is more than D, and the number of capital entities is more than lowercase entities, we adopt entity substitution for data augmentation. The sentences containing only D are randomly replaced with arbitrary D. Then we obtained 1,910 sentences containing only the D entities and the total number of sentences are 7,618 currently. Finally, the D in the 7,618 sentences are randomly replaced with other lowercase D, and the M are randomly replaced with other lowercase M. As a result, the dataset including 15,236 sentences is obtained, called MDdata1. MDdata2 is composed of 58,464 sentences divided by the experimental part of 1,226 papers of the PAKDD conferences (2009-2019), which is used to analyze the development of M.

### B. Experiment setup

MDdata1 is split according to the ratio that training: test: cross-validation is 7.5: 1: 1.5. The specific experiment settings are as follows. For each input sentence, the maximum (character level) length is set to 600, and the batch size is set to 16. The dimension of the rule embedding is 40, and the character embedding dimension is 200. CNN uses 30 convolution kernels with 1*1 and the convolution stride is 1*2. After convolution with the Relu function, max-pooling is used. BiLSTM has two layers, and each layer has 240 hidden units. In the attention layer, the dimension of the involved matrix **W** is 510*480. We implement our model by using Tensorflow 2.0. To evaluate the performance of MDER, we employ accuracy, recall and F1-score. After the model training is completed, MDER is applied to the MDdata2 for predicting M and D entities.

### C. Results

TABLE I. RESULTS ON MDDATA1

| Model | Accuracy | Recall | F1-score |
|---|---|---|---|
| MDER | 0.906 | 0.791 | 0.845 |
| without rule+CNN | 0.813 | 0.698 | 0.751 |

The experimental results on MDdata1 are shown in Table I. MDER without rule+CNN is a baseline which has BiLSTM+Attention+CRF structure, which is the most widely used sequence labeling network. MDER is significantly better than the baseline model. The accuracy, recall, and F1-score of MDER are higher than those without CNN or rule-based embedding. The results show that rule embedding can reduce the learning burden of the model and help model learn. CNN captures structural information according to the current context, which is a useful supplement to BiLSTM.

Applying the trained MDER on MDdata2, we obtain the M and D mentioned in papers. The complex network is built for the extracted M for each year. An edge represents two methods that appear in one paper, and the edge weight is the number of papers containing the two methods. Part of the results are shown in Fig. 2-4.

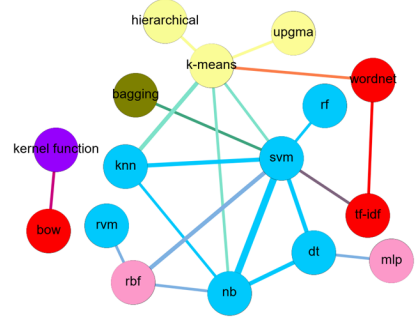

Fig. 2. The method network with edge weights >2 in 2009.

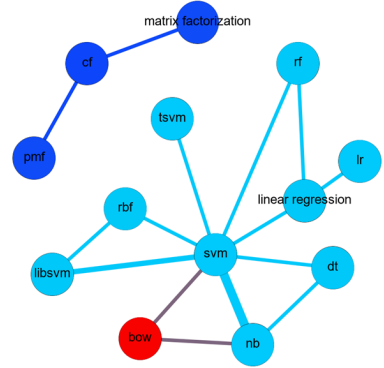

Fig. 3. The method network with edge weights >2 in 2014.

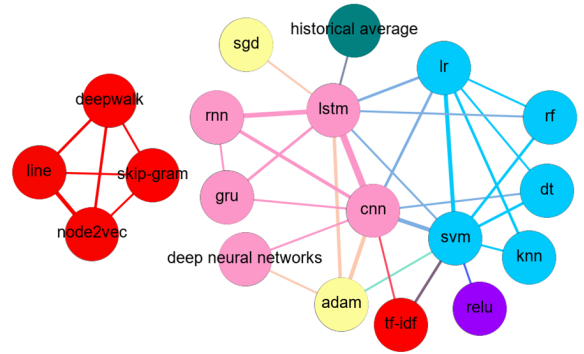

Fig. 4. The method network with edge weights >2 in 2019.

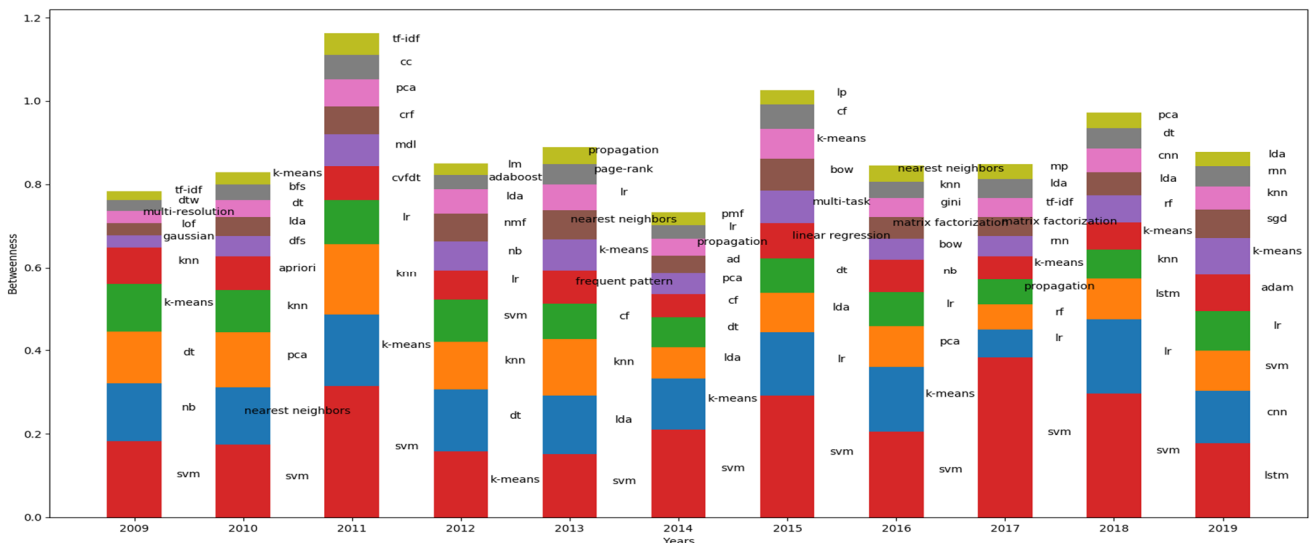

Fig. 5 The histogram of the betweenness centrality of top ten methods from 2009 to 2019.

Fig. 2 presents that the shallow classification algorithms (the blue nodes) in 2009, such as support vector machine (SVM), naive Bayes (NB), decision tree (DT), k-nearest neighbor (KNN) and Random Forest (RF), often appear as comparison methods in many papers. The clustering algorithms (the yellow nodes) also receive a lot of attention, and researchers prefer to use K-means and Hierarchical Clustering methods. TF-IDF and bag-of-words (BOW) are commonly used together as text representation methods (the red nodes). Fig. 3 demonstrates that the shallow classification algorithms are still widely used in 2014, such as SVM, NB, DT, LR, etc., and SVM derives a variety of variants as comparison models, such as LIBSVM and TSVM. Recommendation methods (the dark blue nodes) are also popular, like Probabilistic Matrix Factorization (PMF), Collaborative Filtering (CF). Fig. 4 shows that the deep learning models (the pink nodes) dominate the landscape in 2019. For example, Convolutional Neural Network (CNN), Recurrent Neural Network (RNN), Long Short-Term Memory (LSTM) and Gated Recurrent Unit (GRU) are co-occurrence usually. Some text representation methods (the red nodes) such as node2vec and skip-gram are also popular. Meanwhile, the shallow machine learning models, such as SVM, KNN, DT and Logistic Regression (LR) are still being used.

Fig. 5 presents the statistics of the top ten nodes in the betweenness centrality of complex network for each year. The betweenness centrality is an indicator that describes the importance of a node by the number of shortest paths passing through. The larger the value is, the more important the node is in a network. Furthermore, SVM has the largest betweenness centrality in eight years, indicating that it is closely related to other nodes. It is very likely that it is used as a baseline comparison model in many studies. From 2009 to 2016, the shallow machine learning methods dominate the landscape, but in 2017, deep learning become popular. LSTM rise from the third position in 2018 to the first in 2019 with CNN following in the second position. The deep learning models are gradually dominating the landscape in the AI field.

IV. CONCLUSION

MDER achieves good performance on extracting entities M and D in machine learning papers, and we find some useful insights by the visualization results. From 2009 to 2016, the traditional machine learning methods dominate the landscape, but in the 2017, deep learning become popular. In the future, we will analyze the relationship among datasets, and mine tens of thousands of papers in the AI field to provide researchers with more information about the literature.